\documentclass{article}
\usepackage{spconf,amsmath,graphicx,hyperref}
\usepackage{multirow}
\usepackage[dvipsnames]{xcolor}
\usepackage{kotex}
\usepackage{blindtext}
\usepackage{booktabs}
\usepackage{amssymb}
\usepackage{subcaption}
\usepackage{xcolor}
\usepackage{wrapfig}
\usepackage{makecell}
\usepackage{graphicx}
\usepackage{tikz}
\usepackage{pifont}
\usepackage{array}
\usepackage{rotating}
\usepackage{tabularx}
\usepackage{algorithm}
\usepackage{algpseudocode} 
\algrenewcommand\algorithmicrequire{\textbf{Input:}}
\algrenewcommand\algorithmicensure{\textbf{Output:}}

\usetikzlibrary{arrows.meta,calc}

\usepackage{tcolorbox}
\tcbuselibrary{listings,skins,breakable}
\usepackage{listings,xcolor}
\lstdefinestyle{jsonstyle}{
  basicstyle=\ttfamily\small,
  numbers=none,
  frame=single, rulecolor=\color{black!20},
  showstringspaces=false, breaklines=true, columns=fullflexible
}

\usepackage{array,booktabs,makecell,xcolor,graphicx}

\newcommand{\cmark}{\textcolor{PineGreen}{\ding{51}}}%
\newcommand{\xmark}{\textcolor{red}{\ding{55}}}%

\newcommand{\eg}{\textit{e.g.~}}

\newcommand{\tref}[1]{Table~\ref{#1}}
\newcommand{\fref}[1]{Fig.~\ref{#1}}

\title{Compositional Image Synthesis with Inference-Time Scaling}
\name{Minsuk Ji$^*$, Sanghyeok Lee$^*$, and Namhyuk Ahn\thanks{$^*$ indicates equal contribution.}}
\address{Inha University}

\begin{document}

\maketitle

\begin{abstract}
Despite their impressive realism, modern text-to-image models still struggle with compositionality, often failing to render accurate object counts, attributes, and spatial relations.
To address this challenge, we present a training-free framework that combines an object-centric approach with self-refinement to improve layout faithfulness while preserving aesthetic quality.
Specifically, we leverage large language models (LLMs) to synthesize explicit layouts from input prompts, and we inject these layouts into the image generation process, where a object-centric vision-language model (VLM) judge re-ranks multiple candidates to select the most prompt-aligned outcome iteratively. By unifying explicit layout-grounding with self-refine-based inference-time scaling, our framework achieves stronger scene alignment with prompts compared to recent text-to-image models. The code is available at \url{https://minsuk-ji.github.io/ReFocus/}.
\end{abstract}

\begin{keywords}
text-to-image synthesis, inference-time-scaling, object-centric
\end{keywords}

\section{Introduction}

Text-to-image (T2I) diffusion models now deliver striking realism and diversity from textual prompts~\cite{rombach2022ldm,sdxl,ramesh2022unclip,flux2024}, yet they still struggle with \emph{compositionality}: the precise rendering of object counts, attributes, and spatial relations~\cite{geneval}.
For example, a prompt ``a photo of four giraffes" often yields the wrong number of animals.
Similarly, relational prompts such as ``a photo of a chair left of a zebra" can lead to the spatial inconsistency.
Such limitation expose a persistent gap between user intent and model output~\cite{geneval}.

To address these issues, recent studies have investigated layout-grounding image generation~\cite{gligen,controlnet,migc}.
However, these methods face two key challenges.
First, they require users to provide both a text prompt and a layout (\eg bounding boxes), which is a cumbersome task.
Second, the stochastic nature of diffusion often leads to inconsistent fidelity.
As a result, many users generate multiple samples to obtain a satisfactory outcome, which further reduces usability.
Our objective is therefore to develop a framework that specifies explicit compositional structure while ensuring high-fidelity rendering in a user-friendly manner.

Recently, diffusion models have embraced inference-time scaling to enhance generation quality~\cite{ma2025inference,bai2024zigzag}. Strategies like Best-of-N~\cite{ma2025inference}  generate multiple samples and select the best one using a VLM judge. While this improves scene-level alignment, it fails to enforce fine-grained fidelity during the generation process. Iterative refinement methods~\cite{reflectdit,reflectionflow} address this by progressively revising outputs via in-context reflection. However, these approaches remain limited as they are not fully object-centric, often overlooking local details—and typically require computationally expensive reflection-tuning.

\begin{table}[t]
\centering
\caption{Summary of representative text-to-image synthesis methods categorized by object-centric, inference-time scaling, self-refine, and training-free properties.}
\vspace{-0.5em}
\label{tab:feature_comparison}
\setlength{\tabcolsep}{2.5pt}
\footnotesize
\renewcommand{\arraystretch}{1.1}
\begin{tabular}{l|c c c c}
\hline
\multirow{2}{*}{Method Group} & Object- & Inference-Time & Self- & Training- \\
& Centric & Scaling & Refine & Free \\
\hline\hline
\scriptsize{SD1.5~\cite{rombach2022ldm}, SDXL~\cite{sdxl}, FLUX~\cite{flux2024}} & \xmark & \xmark & \xmark & \cmark \\
\scriptsize{GLIGEN~\cite{gligen}, ControlNet~\cite{controlnet}} & \cmark & \xmark & \xmark & \xmark \\
\scriptsize{Best-of-$N$~\cite{ma2025inference}, Z-Sampling~\cite{bai2024zigzag}} & \xmark & \cmark & \xmark & \cmark \\
\scriptsize{Reflect-DiT~\cite{reflectdit}}     & \xmark & \cmark & \cmark & \xmark  \\
\hline
\texttt{ReFocus} (ours) & \textbf{\cmark} & \textbf{\cmark} & \textbf{\cmark} & \textbf{\cmark} \\
\hline
\end{tabular}
\vspace{-2em}
\end{table}


\begin{figure*}[t]
\centering
\includegraphics[width=\linewidth, height=5.5cm]{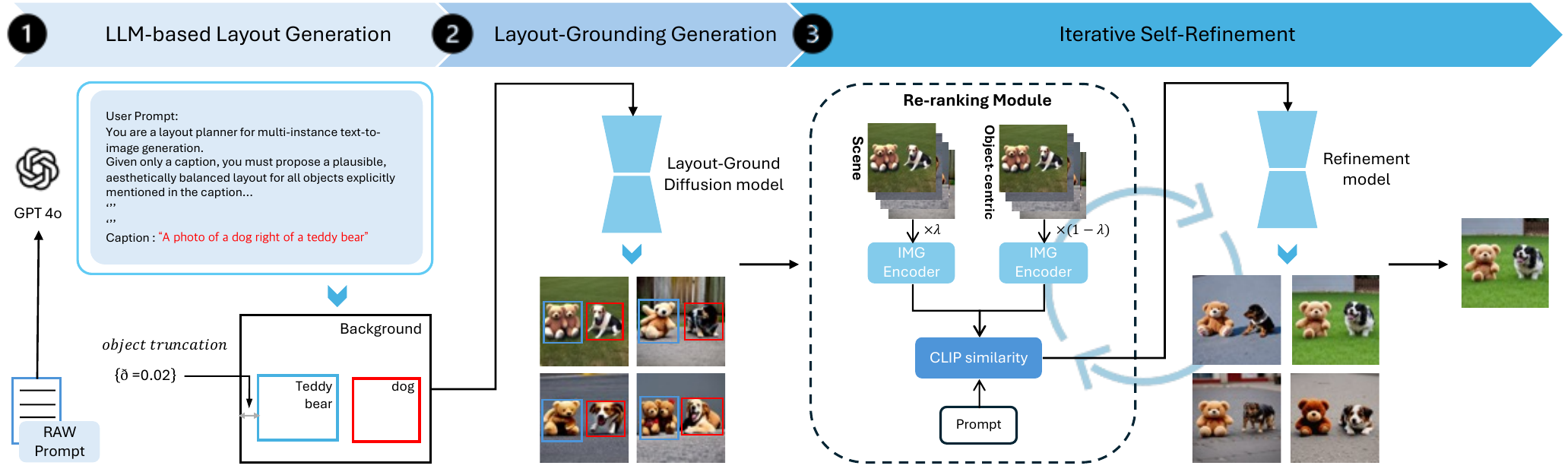}
\caption{\textbf{Overview of \texttt{ReFocus}.} (1) \emph{LLM-based Layout Generation:} The prompt is mapped to an explicit box layout $L$ and lightly regularized (2\% border margin, $\delta{=}0.02$) to avoid truncation. (2) \emph{Layout-Grounding Generation:} a diffusion model conditioned on $L$ samples $N$ drafts. (3) \emph{Iterative self-refinement:} a hybrid re-ranking module, weighted by $\lambda$ selects best candidate and the refinement model iteratively refines and re-ranked candidates until a prompt consistent image is produced.}
\label{fig:pipeline_overview}
\vspace{-1em}
\end{figure*}

To this end, we propose \texttt{ReFocus}, a training-free framework that integrates inference-time refinement with an object-focused perspective. Inspired by the recent observations that LLMs are strong at layout understanding~\cite{layoutgpt}, we leverage LLMs to automatically generate layout groundings from input prompts, removing the need for users to provide layouts manually. These layouts are injected into the generation process and iteratively refined to enhance both faithfulness and fidelity. Specifically, an object-focused VLMs judge evaluates candidate scenes, re-ranks them to identify the current optimal sample, and further guides self-refinement to revise the generation in an iterative loop.

As summarized in \tref{tab:feature_comparison}, our framework integrates all key features necessary to improve the usability of text-to-image synthesis. In particular, unlike Best-of-$N$~\cite{ma2025inference}, which applies inference-time scaling but lacks refinement, our approach incorporates object-centric and self-refinement. Similarly, unlike Reflect-DiT~\cite{reflectdit}, which refine generation during inference but remain scene-level and require additional training, our method is both object-centric and training-free. With this simple yet powerful design, our framework achieves strong performance on challenging image synthesis benchmarks such as GenEval~\cite{geneval} and HPS~\cite{hpsv2}.
\section{Proposed Method}

Our primary objective is to generate high-fidelity image from a complex compositional prompt $P$. The goal is to produce outputs that are both semantically and structurally faithful to the prompt, while keeping the overall process user-friendly and training-free. To achieve this, we introduce \texttt{ReFocus}, a novel framework that synergistically integrates the strengths of prior approaches (in \tref{tab:feature_comparison}) without requiring additional training. As illustrated in Fig.~\ref{fig:pipeline_overview}, our framework proceeds in several phases: it first establishes a compositionally accurate basis through explicit layout grounding, and then progressively enhances aesthetic quality through hierarchical refinement and re-ranking based on inference-time scaling.

\begin{table*}[t]
\centering
\caption{Quantitative comparison on the GenEval benchmark dataset~\cite{geneval}. Higher score indicates better performance.}
\label{tab:comparison}
\footnotesize
\vspace{-0.9em}
\setlength\tabcolsep{11pt}
\begin{tabular}{l|c|cccccc}
\hline
Method & Average & Single & Two & Counting & Colors & Position & Attribution \\
\hline\hline
SD1.5~\cite{rombach2022ldm} & 0.43 & 0.97 & 0.38 & 0.35 & 0.76 & 0.04 & 0.06 \\
SDXL~\cite{sdxl} & 0.55 & 0.98 & 0.74 & 0.39 & 0.85 & 0.15 & 0.23 \\
SDXL~\cite{sdxl} + GLIGEN~\cite{gligen} & 0.65 & 0.95 & 0.73 & 0.70 & 0.72 & 0.56 & 0.23 \\
\hline
FLUX.1-dev~\cite{flux2024} & 0.68 & 0.99 & 0.85 & 0.74 & 0.79 & 0.21 & 0.48 \\
SD3~\cite{esser2024scaling} & 0.74 & 0.99 & 0.94 & 0.72 & \textbf{0.89} & 0.33 & \textbf{0.60} \\
DALL-E~3~\cite{betker2023dalle3} & 0.67 & 0.96 & 0.87 & 0.47 & 0.83 & 0.43 & 0.45 \\
\hline
SD1.5~\cite{rombach2022ldm} + Best-of-$N$~\cite{ma2025inference} ($N=4$) & 0.51 & 0.98 & 0.59 & 0.46 & 0.85 & 0.09 & 0.12 \\
SDXL~\cite{sdxl} + Best-of-$N$~\cite{ma2025inference} ($N=4$) & 0.61 & 0.98 & 0.82 & 0.61 & \textbf{0.89} & 0.09 & 0.26 \\
Sana-1.0-1.6B~\cite{xie2025sana} + Best-of-$N$~\cite{ma2025inference} ($N=20$) & 0.75 & 0.99 & 0.87 & 0.73 & 0.88 & 0.54 & 0.55 \\
SDXL + Z-Sampling~\cite{bai2024zigzag} & 0.57 & \textbf{1.00} & 0.74 & 0.46 & 0.87 & 0.10 & 0.24 \\
Reflect-DiT~\cite{reflectdit} + Best-of-$N$~\cite{ma2025inference} ($N=20$) & 0.81 & 0.98 & \textbf{0.96} & 0.80 & 0.88 & 0.66 & 0.60 \\
\hline
\texttt{ReFocus} ($N=4$) (ours) & \textbf{0.84} & 0.99 & 0.92 & \textbf{0.82} & 0.86 & \textbf{0.81} & \textbf{0.60} \\
\hline
\end{tabular}
\vspace{-1.8em}
\end{table*}

\subsection{Phase 1: LLM-based Layout Generation}
Given an unstructured prompt $P$, the initial phase translates it into an explicit layout representation $L=\{(l_i,s_i)\}_{i=1}^{k}$, where each object label $l_i$ is paired with its corresponding spatial layout $s_i$, and $k$ denotes the number of objects. Recent studies have shown that LLMs are remarkably capable of representing spatial layouts~\cite{layoutgpt}. Motivated by this, we employ an LLM to parse the input prompt and return a layout $L$, formulated as $L = f_{\mathrm{LLM}}(P)$, where $f_{\mathrm{LLM}}$ denotes the layout parser. Each layout $s_i$ is represented by normalized coordinates $s_i \in [0,1]^4$ in $(x_{\min}, y_{\min}, x_{\max}, y_{\max})$ format. Unlike prior methods that rely on manual layout annotation~\cite{gligen,controlnet}, this automated process supplies explicit spatial guidance for the subsequent layout-grounded generation.

To improve robustness and accommodate complex spatial relationships, we refine our instruction prompts using an adaptive margin strategy grounded in the model architecture:
(1) We shrink layout boxes by a margin $\delta \in [0.02, 0.04]$. This range is calibrated to align with the Latent Diffusion architecture; considering the $1/8$ downsampling factor (e.g., $512 \times 512 \rightarrow 64 \times 64$), a margin of 0.02 corresponds to approximately 1.28 pixels in the latent space. This acts as a critical boundary to prevent "concept bleeding" between adjacent objects.
(2) Moving beyond a rigid non-overlap constraint, we employ a relation-aware adjustment. The LLM parses spatial descriptions to distinguish between independent and interacting objects. We relax the margin constraints for objects with explicit depth dependencies (e.g., "behind") to allow natural occlusion, while enforcing stricter boundaries for spatially distinct objects to maintain generation fidelity.

\subsection{Phase 2: Layout-Grounding Initial Generation}
Given the input prompt $P$ and the LLM-generated layout $L$, we perform layout-grounding image generation to obtain an initial draft. Using a layout-conditioned diffusion model $G$, we synthesize a set of $N$ draft images $\mathcal{I}_{\mathrm{draft}}=\{I_1,\dots,I_N\}$ from independent standard Gaussian noise vectors $z_i$:
\begin{equation}
I_i = G(P,L,z_i), \quad i=1,\ldots,N.
\label{eq:initial_generation}
\end{equation}

Unlike standard text-to-image models~\cite{rombach2022ldm,sdxl} that often misplace objects or fail to capture relationships, this phase leverages the layout to impose a coarse compositional structure from the outset from an object-centric viewpoint. In addition, unlike existing layout-grounding image synthesis models~\cite{gligen,controlnet,migc}, we do not treat this as the final output. Instead, we generate $N$ diverse drafts that serve as the basis for subsequent self-refinement. This design enables our framework to automatically produce high-quality outputs for complex image generation tasks without requiring users to engage in cumbersome manual trial-and-error.

\begin{figure}[t]
    \centering
    \vspace{-0.1em}
    \includegraphics[width=\linewidth]{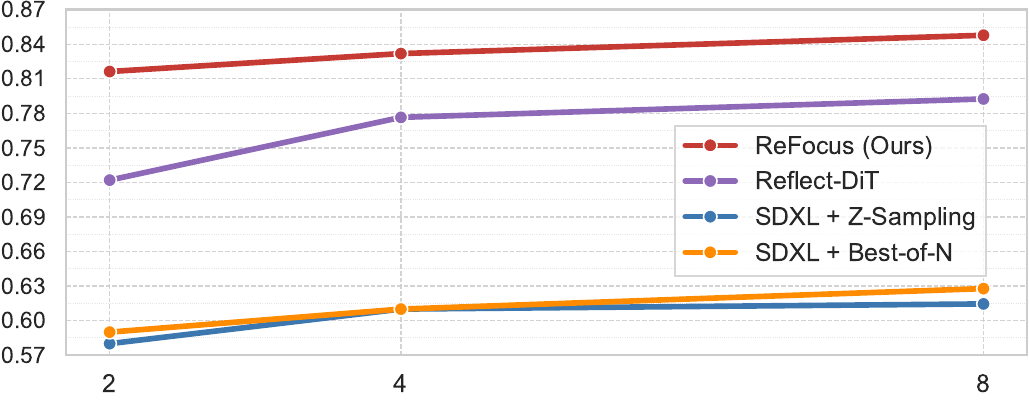}
    \vspace{-1.8em}
    \caption{Average GenEval~\cite{geneval} score as the number of samples per prompt ($N$ in Best-of-$N$) increases.}
    \label{fig:overall_vs_n}
    \vspace{-1.2em}
\end{figure}

\subsection{Phase 3: Iterative Self-Refinement}

While the initial draft $\mathcal{I}_{\mathrm{draft}}$ is structurally aligned due to object-centric synthesis, they often lack photorealistic detail. Moreover, since the backbone diffusion model~\cite{sdxl} has difficulty handling overlapping objects, the resulting images can be of poor quality when such cases arise. To address these, this phase introduces an iterative process of validating and refining the drafts, progressively improving image fidelity.

\smallskip
\noindent\textbf{Preference Re-ranking.}
Given the draft set $\mathcal{I}_{\mathrm{draft}}$, we perform preference re-ranking to identify the most promising candidates. Unlike prior approaches that rely solely on scene-level CLIP similarity~\cite{ma2025inference}, we introduce a hybrid evaluation that integrates both scene-level and object-level judgments.
The scene-level score $S_{\mathrm{scene}}$ measures holistic alignment between the prompt $P$ and the generated image using standard CLIP~\cite{clip} similarity, which captures global fidelity such as spatial relations and overall semantics as in below.
\begin{equation}
S_{\mathrm{scene}}(I,P)\equiv S_{\mathrm{CLIP}}(I,P).
\label{eq:global_score}
\end{equation}

In parallel, we compute an object-level preference $S_{\mathrm{object}}$ by first extracting object descriptions from $P$, and then cropping object regions from each draft image according to the LLM-generated layout $L$. We then compute the average CLIP similarity across these cropped regions, yielding an object-centric score. This local score effectively audits each object’s presence and identity within the scene.
\begin{equation}
S_{\mathrm{object}}(I,P,L)\equiv \tfrac{1}{k}\sum_{i=1}^{k} S_{\mathrm{CLIP}}(\mathrm{Crop}(I,l_i),\, P_i),
\label{eq:local_score}
\end{equation}
where $P_i$ indicate parsed object-wise description from $P$.
The overall preference score is defined as $S = \lambda\,S_{\mathrm{scene}} + (1-\lambda)\,S_{\mathrm{object}}$, where $\lambda$ is a hyperparameter that balances scene-level and object-level alignment.
We re-rank the drafts based on $S(I, P, L)$ and retain the top-$K$ candidates for refinement. This hybrid scoring provides a more object-centric preference evaluation compared to prior scene-only verification.

\begin{figure}[t]
    \centering
    \includegraphics[width=\linewidth]{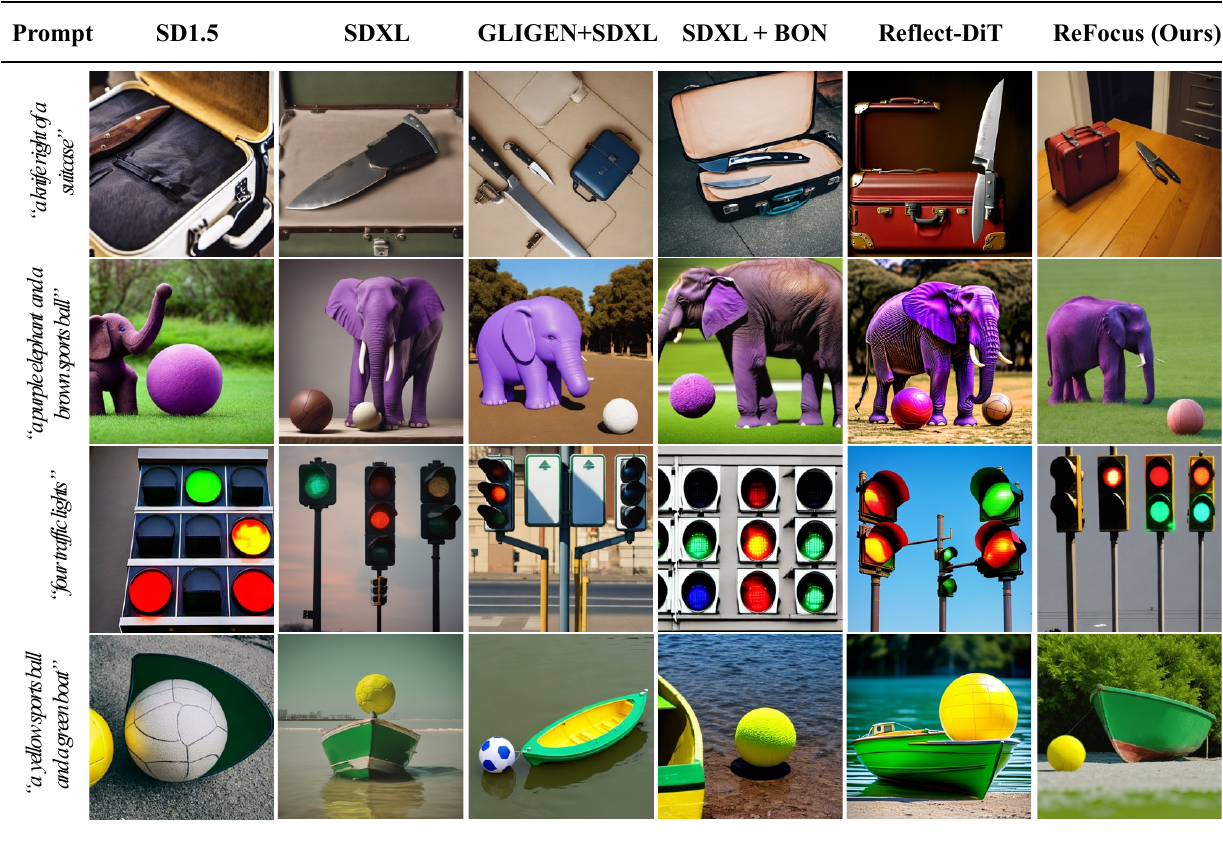}
    \vspace{-2em}
    \caption{Visual comparison with prior text-to-image models~\cite{rombach2022ldm,sdxl}, a layout-grounding method~\cite{gligen}, and inference-time scaling approaches~\cite{ma2025inference,reflectdit}.}
    \label{fig:qual}
    \vspace{-1em}
\end{figure}

\smallskip
\noindent\textbf{Refinement.}
Given the top-$K$ ranked image set $\mathcal{I}_{\mathrm{top-k}}$, we employ a lightweight refinement model $G_{\mathrm{refine}}$ to enhance visual quality while preserving compositional structure. Following the best-of-$N$ principle~\cite{ma2025inference}, we generate $M$ variants by applying independent noise and partial denoising~\cite{meng2021sdedit}:
\begin{equation}
I_i = G_{\mathrm{refine}}(\mathcal{I}_{\mathrm{top-k}}, z_i;\alpha_{\mathrm{refine}}), \quad \alpha_{\mathrm{refine}} \ll 1,
\end{equation}
where $\alpha_{\mathrm{refine}}$ is a low denoising strength. The refined set is then given by
$\mathcal{I}_{\mathrm{refined}} = \{I_1,\dots,I_M\}$.
This step enriches fine-grained details such as texture, lighting, and shading without altering the geometric arrangement established by the layout grounding. Unlike standard SDEdit-based resampling~\cite{meng2021sdedit}, our refinement is coupled with preference re-ranking so that realism improvements do not come at the expense of compositional faithfulness.

\smallskip
\noindent\textbf{Iterative Process}.
We repeat the preference re-ranking and refinement steps in an iterative loop, forming a self-refining process guided by object-aware judgment. Each iteration progressively improves both scene-level alignment and object-level fidelity. As a result, the model can correct errors such as missing objects or implausible details. This design distinguishes the proposed \texttt{ReFocus} from prior inference-time scaling approaches~\cite{ma2025inference}, which stop after a single best-of-$N$ selection, as well as from reflection-based refinement methods~\cite{reflectdit}, which require additional training. \texttt{ReFocus} remains training-free and achieves stronger prompt fidelity in complex compositional image synthesis.

\section{Experimental Results}

\noindent\textbf{Implementation details.}
We employ MIGC~\cite{migc} as the layout-conditioned diffusion model $G$ in Phase 2.
For Phase 3, we adopt SDXL-Turbo~\cite{sdxlturbo} as the refinement model, chosen for its fast inference, which can minimize latency in the iterative loop.
In Phase 1, we use ChatGPT-4o as the LLM for layout parsing. During Phase 2, each image is generated with 50 sampling steps and a classifier-free guidance~\cite{cfg} scale of 7.5. In Phase 3, we apply a single refinement step with the guidance fixed at 0.0 and the denoising strength set to 0.5.

\smallskip
\noindent\textbf{Baselines.}
We compare with representative diffusion-based text-to-image models~\cite{rombach2022ldm,sdxl}, layout-grounding model~\cite{gligen}, inference-time scaling methods~\cite{ma2025inference,bai2024zigzag} and feedback-based scaling method~\cite{reflectdit}. These serve as strong baselines to evaluate overall performance across categories.

\smallskip
\noindent\textbf{Evaluation.}
We evaluate our model using the GenEval benchmark~\cite{geneval}, which is specifically designed to measure prompt faithfulness in object-focused attributes. We adopt GenEval to assess object-level compositional accuracy. To analyze the impact of iterative refinement of our \texttt{ReFocus}, we additionally employ the Human Preference Score (HPS v2.1)~\cite{hpsv2}, which measures visual quality and human preference.

\begin{table}[t]
\centering
\caption{Ablation study on the effect of inference-time scaling (ITS) and self-refinement (Refine). Higher score is better.}
\vspace{-0.5em}
\label{tab:ablation}
\footnotesize
\setlength{\tabcolsep}{3pt}
\begin{tabular}{l|c|ccccc}
\hline
\multirow{2}{*}{Model} & \multicolumn{1}{c|}{GenEval~\cite{geneval}} & \multicolumn{5}{c}{HPS v2~\cite{hpsv2}} \\
\cline{2-7}
 & Avg. & Avg. & Anime & Art & Painting & Photo \\
\hline\hline
SD1.5~\cite{rombach2022ldm}  & 0.43 & 27.12 & 27.43 & 26.71 & 26.73 & 27.62 \\
SD2.0~\cite{rombach2022ldm}  & 0.51 & 27.17 & 27.48 & 26.89 & 26.86 & 27.46 \\
\hline
Phase 1 & 0.78 & 26.74 & 26.56 & 26.04 & 26.23 & 28.15 \\
+ ITS (BoN~\cite{ma2025inference}) & 0.80 & 27.17 & 27.20 & 26.47 & 26.71 & 28.29 \\
+ Refine (Round 1) & 0.80 & 28.29 & 28.60 & 27.65 & 28.00 & \textbf{28.91} \\
+ Refine (Round 2) & \textbf{0.84} & \textbf{28.32} & \textbf{28.85} & \textbf{27.97} & \textbf{29.06} & 28.55\\
\hline
\end{tabular}
\vspace{-1.5em}
\end{table}

\subsection{Model Comparison}
\noindent\textbf{Quantitative results.}
As in \tref{tab:comparison}, \texttt{ReFocus} achieves the highest average score on GenEval~\cite{geneval} and shows consistent gains across object-focused categories. For example, compared with SDXL~\cite{sdxl} baseline, \texttt{ReFocus} improves position by +0.66, counting by +0.43. Compared with Reflect-DiT~\cite{reflectdit}, which also combines inference-time scaling with self-revision, our method is training-free, requiring no feedback collection or alignment tuning, and it achieves better performance with fewer inference samples ($N$). Furthermore, even against GLIGEN~\cite{gligen}, which explicitly conditions on layouts, our method yields higher accuracy in position. We attribute this to the synergy between object-centric grounding and inference-time scaling.

\smallskip
\noindent\textbf{Visual results.}
In \fref{fig:qual}, our method preserves prompt-aligned scene structure while improving perceptual quality. Prompts such as ``a purple elephant and a brown sports ball" and ``four traffic signs" demonstrate accurate object counts, faithful colors, and correct relative positions, whereas prior methods often miss objects or distort spatial relations.

\smallskip
\noindent\textbf{Increasing \# samples.}
\fref{fig:overall_vs_n} further examines performance as the number of inference-time samples increases. Our curve consistently stays above competing methods, and the gap remains as $N$ grows. This result suggests that explicit layout grounding combined with object-centric refinement provides a scalable solution rather than a one-off heuristic.

\subsection{Model Analysis}
\begin{figure}[t]
    \centering
    \includegraphics[width=\linewidth]{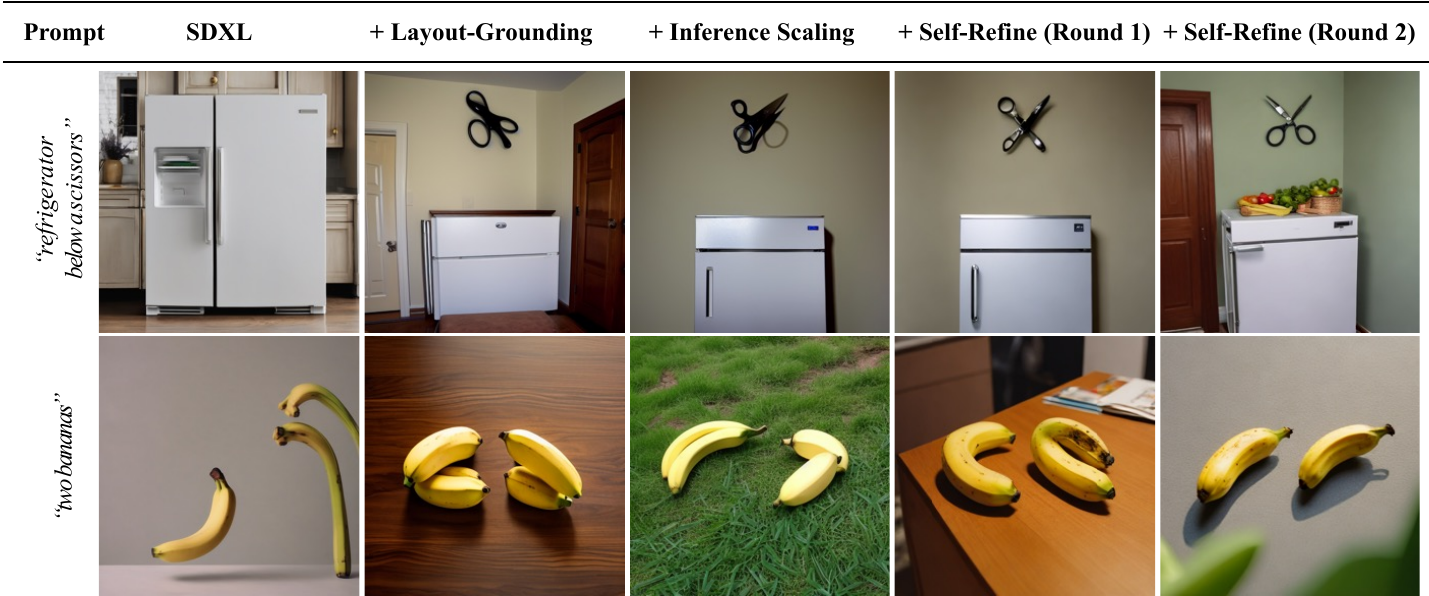}
    \vspace{-1.5em}
    \caption{Visual comparison of our proposed mechanism. Here, inference scaling refers to the naive Best-of-$N$ strategy.}
    \label{fig:ablation}
    \vspace{-1.5em}
\end{figure}

In \tref{tab:ablation}, Phase~1 (layout grounding only) secures coarse compositional structure (moderate GenEval), but fine object details and overall realism remain limited (low HPS v2). The inference-time scaling provides only modest improvements, particularly on HPS v2. In contrast, the self-refinement loop markedly improves performance on HPS v2, with substantial gains in perceptual quality
In \fref{fig:ablation}, layout grounding establishes a reliable geometric basis; preference re-ranking selects the most prompt-aligned candidate; and refinement sharpens texture, lighting, and boundaries while preserving geometry. This sequence reduces missing objects and corrects implausible local details without drifting from the input prompt.

\section{Conclusion}
In this paper, we have introduced \texttt{ReFocus}, a training-free framework for compositional text-to-image synthesis that integrates an object-centric approach with inference-time scaling-based iterative self-refinement. This design enables our method to generate images that are both visually appealing and strongly aligned with the input prompt. The simplicity and effectiveness of our framework make it a practical step toward reliable and user-friendly text-to-image generation.

\noindent\textbf{Acknowledge.} This work was supported by Institute of Information \& communications Technology Planning \& Evaluation (IITP) under the Leading Generative AI Human Resources Development (IITP-2026-RS-2024-00360227) grant funded by the Korea government (MSIT).

\bibliographystyle{IEEEbib}
\bibliography{main}


\end{document}